\documentclass[10pt, conference, compsocconf]{IEEEtran}
\usepackage{algorithm}
\usepackage{algorithmicx, algpseudocode}
\usepackage{multirow}
\usepackage{amsmath}
\usepackage{amsfonts}
\usepackage{graphics}
\usepackage{graphicx}
\usepackage{subfigure}
\usepackage{threeparttable}
\usepackage{url}

\begin{document}
%
% paper title
% can use linebreaks \\ within to get better formatting as desired
\title{LSBN: A Large-Scale Bayesian Structure Learning Framework for Model Averaging}

\author{\IEEEauthorblockN{Yang Lu\IEEEauthorrefmark{1},
Mengying Wang\IEEEauthorrefmark{2},
Menglu Li\IEEEauthorrefmark{1},
Qili Zhu\IEEEauthorrefmark{3} and
Bo Yuan\IEEEauthorrefmark{1}}
\IEEEauthorblockA{\IEEEauthorrefmark{1}Department of Computer Science\\
Shanghai Jiao Tong University,
Shanghai, China\\ Email: \{luyang0415, iby07, boyuan\}@sjtu.edu.cn}
\IEEEauthorblockA{\IEEEauthorrefmark{2}Software Engineering Institute\\
East China Normal University,
Shanghai, China\\ Email: mywang@sei.ecnu.edu.cn}
\IEEEauthorblockA{\IEEEauthorrefmark{1}Department of Computer Science\\
Shanghai Jiao Tong University,
Shanghai, China\\ Email: kzhu@cs.sjtu.edu.cn}}

% make the title area
\maketitle

\begin{abstract}
%\boldmath

The motivation for this paper is to apply Bayesian structure learning using Model Averaging in large-scale networks. Currently, Bayesian model averaging algorithm is applicable to networks with only tens of variables, restrained by its super-exponential complexity. We present a novel framework, called \emph{LSBN}(Large-Scale Bayesian Network), making it possible to handle networks with infinite size by following the principle of divide-and-conquer.

The method of \emph{LSBN} comprises three steps. In general, \emph{LSBN} first performs the partition by using a second-order partition strategy, which achieves more robust results. \emph{LSBN} conducts sampling and structure learning within each overlapping community after the community is isolated from other variables by Markov Blanket. Finally \emph{LSBN} employs an efficient algorithm, to merge structures of overlapping communities into a whole.

In comparison with other four state-of-art large-scale network structure learning algorithms such as ARACNE, PC, Greedy Search and MMHC, \emph{LSBN} shows comparable results in five common benchmark datasets, evaluated by precision, recall and f-score. What's more, \emph{LSBN} makes it possible to learn large-scale Bayesian structure by Model Averaging which used to be intractable.

In summary, \emph{LSBN} provides an scalable and parallel framework for the reconstruction of network structures. Besides, the complete information of overlapping communities serves as the byproduct, which could be used to mine meaningful clusters in biological networks, such as protein-protein-interaction network or gene regulatory network, as well as in social network.

\end{abstract}
% IEEEtran.cls defaults to using nonbold math in the Abstract.
% This preserves the distinction between vectors and scalars. However,
% if the conference you are submitting to favors bold math in the abstract,
% then you can use LaTeX's standard command \boldmath at the very start
% of the abstract to achieve this. Many IEEE journals/conferences frown on
% math in the abstract anyway.

% no keywords

\IEEEpeerreviewmaketitle

\section{Introduction}

Structure learning from sparse data serves as a central problem in a variety of research area, for it uncovers underlying relationships, dependencies among variables, and more importantly, brings forth a structured, easily-understood model for further prediction and inference. As a major structure learning approach, a Bayesian network describes a probabilistic graphical model by representing a set of random variables and conditional dependencies via a directed acyclic graph (DAG). What's more, a Bayesian network provides a very flexible framework to fuse different types of data and prior knowledge together to derive a synthesized network.

To achieve more robust and proper results in Bayesian structure learning, it is preferable to integrate over all possible structure models by using Bayesian model averaging. However, with the number of network variables growing, the enumeration of all possible structures becomes intractable and impractical, for there are overall $O(n!2^{\binom{n}{2}})$ possible structures given \emph{n} network variables\cite{Robinson1973}. In short, structure learning by using model averaging is NP-hard even when the maximum parents number of network variable is bound to certain constant value \emph{k}\cite{DBLP:journals/ml/HeckermanGC95}. However, in real applications, ranging from casuality network to Protein-Protein-Interaction network, the scales are much larger than traditional structure learning by using model averaging could support.

A very natural and logical attempt to scale the Bayesian structure learning beyond its limitation of variable numbers is to partition the variables into multiples groups, thus employing the method separately and efficiently. Manual partition is one option\cite{DBLP:journals/jmlr/KoivistoS04}, yet subjective factors would inevitably play a nontrivial role and possibly influence the ultimate result. Another widely-applied approach involves prior knowledge\cite{V.K.Mansinghka:2006:8d66d}, where domain knowledge is exploited to distinguish closely related variables thus guide the partition. For example, a common application under guidance of prior knowledge is Gene Regulatory Network(GRN) inference from gene expression data. In this case, cluster analysis is frequently applied to find similar functional groups, based on the assumption that genes presented by similar expression patterns tend to be co-regulated or interact\cite{geneRegulatoryNetReview}.

Unfortunately, the partition strategy is confronted with three fundamental limitations. The first problem lies in the lack of prior knowledge in most cases. It is neither practical nor tractable to collect prior knowledge for a special purpose beforehand. The second problem is that even there does exist prior knowledge, it remains challenges to quantify the knowledge as the network prior distribution. For example, if the prior distribution is assigned with higher value, significant bias could be resulted towards the prior knowledge, leading to unwarranted structures without paying sufficient attention to data. Or the prior distribution is insignificant, it won't do help to improve learning results. The third problem arises when we attempt to obtain prior knowledge directly from data by using statistical measurements, such as correlation coefficients\cite{Stuart+al:Science03}, mutual information\cite{DBLP:journals/bmcbi/MargolinNBWSFC06}. However, these measurements are limited to pairwise information so that different measurements would inevitably lead to different partition results.

In this paper, we propose a novel framework \emph{LSBN} (\textbf{L}arge-\textbf{S}cale \textbf{B}ayesian \textbf{N}etwork) to learn Bayesian structure for sufficiently large networks. The basic idea of our framework is motivated by the philosophy of divide-and-conquer. Specifically, \emph{LSBN} recursively partitions network variables into multiple communities with much smaller sizes, learns intra-community variables respectively before merge them altogether again. Our contributions lie in three aspects: \\

$\bullet$ We propose a robust partition algorithm, called '\emph{ROPART}', to segment large-scale network variables into multiple overlapping communities with much smaller sizes. According to the traditional graph clustering problem, whether variables are allocated into the same group depends on their edge weight, denoting the interrelated closeness among each other. Therefore, how to measure edge weights among variables is a challenge. Common measurements, such as mutual information\cite{DBLP:journals/bmcbi/MargolinNBWSFC06}, pearson coefficient\cite{Stuart+al:Science03}, show different partition results given the same data. No one dominate the other, for each one performs excellent in some datasets and dissatisfactory in others. So \emph{ROPART} introduces a second-order partition strategy, to overcome this shortcoming in a robust way. (Section \ref{partition})\\

$\bullet$ We propose a sampling strategy to generate smaller sub-communities when current community is still too large to perform practical Bayesian structure learning. Also, we figure out how to isolate the dependencies of intra-community variables from those outside the community. The isolation makes intra-community structure learning unbiased and credible. What's more, we categorize and analyze primary types of error edges from structure learning, and apply a uniform strategy to resolve the problem with satisfactory results. (Section \ref{learn})\\

$\bullet$ We propose an efficient algorithm, called '\emph{MERGENCE}' to merge the intra-community results into a whole. \emph{MERGENCE} involves seeking an efficient mergence order, and resolves conflicts during the process of mergence. (Section \ref{merge}) \\

We benchmark evaluation of \emph{LSBN} on five well-known datasets, in comparison with four state-of-art structure learning algorithms. The compared results reveal that \emph{LSBN} achieves comparable results to state-of-art structure learning algorithms, meanwhile, \emph{LSBN} makes it possible to learn Bayesian structure by Model Averaging which used to be intractable in large-scale network.

\section{Related Work}
\label{relate}

The problem of static Bayesian Structure Learning is well-studied. Exclusive of Bayesian Model Averaging, the four major approaches are information-theoretic, constrain-based, score-and-search and hybrid. And several representative algorithms would be included in performance evaluation (Section \ref{experiments})

The first major approach to Bayesian Structure Learning is based on Information Theory Models. They weigh network edges by correlation coefficients or statistic scores derived from Mutual Information\cite{geneRegulatoryNetReview}, such as RELNET\cite{relnet}, ARACNE\cite{DBLP:journals/bmcbi/MargolinNBWSFC06} and CLR\cite{clr}. Though most of information-theoretic approaches are subject to unweighted networks, an asymmetric variation of Mutual Information measurement could be employed to obtain directed networks\cite{Rao_usingdirected}. The advantages of Information Theory Models lie in extreme simplicity and low computational cost. However, such models could only take into consideration pairwise relationship rather than multiple variables at the same time. Another drawback is such models usually require plenty of observation data for the sake of accuracy.

The second major approach is constrain-based algorithm. Specifically, constrain-based algorithms use conditional independence (CI) tests to reveal the target DAG, such as PC\cite{Spirtes2000} and RAI\cite{rai}. The drawback of such algorithms is that number of requisite CI tests grows exponentially with the number of variables, so polynomial complexity could only be ensured by imposing the maximum parents number. Besides, such algorithms inevitably miss or wrongly identify V-structures, which would affect the orientation of edges and even subsequent stages.

The third major approach performs a score-and-search strategy. In general, score-and-search algorithms search through structure space guided by a scoring function. One of the most basic score-and-search algorithms is Greedy Search\cite{Brown05acomparison}. Since the size of structure space grows super-exponentially, the search approach would get inevitably trapped into local maximum, even there are many ways to escape, such as random restarts, simulated annealing or search in the space of equivalence classes of DAGs, called PDAGs.

The forth major approach serves as a hybrid approach. Hybrid approaches integrate constrain-based and score-and-search algorithms together. MMHC\cite{Tsamardinos06themax-min} (Max-Min Hill-Climbing) shows superiority to other algorithms by combining local learning, reconstructing the skeleton of a Bayesian network by constrain-based approach, and performing greedy hill-climbing search for edge orientation.

Great amounts of work has been devoted to detecting overlapping communities in large networks. The most popular algorithm is Clique Percolation Method (CPM)\cite{PalEtAl05} which computes all \emph{k} cliques and two variables belong to the same cluster if there exists a path going through \emph{k-1} cliques between them. CPM is implemented by CFinder\cite{cfinder} (http://www.cfinder.org/). Besides, overlapping communities detection could be roughly categorized into threefold: optimization, clustering and partitioning.

One of traditional methods regards overlapping communities detection as an optimization problem, specifically, each community is identified as a subgraph reaching local optimization given quality function $W$, thus detecting overlapping communities becomes finding all locally-optimized subgraphs\cite{Baumes04discoveringhidden}. Furthermore, the optimization could be augmented by combination with spectral mapping and fuzzy clustering\cite{ZhaWanZha07}.

Clustering approaches define clusters as either the set of nodes or the set of edges, and then perform the clustering according to the similarity among nodes\cite{nepusz07b} or edges\cite{YY_LC_nature2010}, respectively. Besides, clustering could be conducted in an agglomerative hierarchy, for example, LinkComm\cite{YY_LC_nature2010} merges groups of edges pairwise in descending order of edge similarity and consequently achieve a dendrogram.

Partitioning approaches transform original graph into a larger graph without overlapping nodes before conduct traditional partitions. Those overlapping nodes are identified and split into multiple copies of themselves beforehand\cite{peacock}. The identification of candidate overlapping nodes is based on \emph{split betweenness}\cite{newman-girvan02}, and the splitting process continues as long as the \emph{split betweenness}
of variables is sufficiently high.

\section{Definitions}

\subsection{DEFINITION 1 (Weight Function)}

Given a generic undirected, unweighted graph $\mathcal{G}=(V,E)$, where $E \in V\times V$. Weight Function \emph{f} maps any edges $e=(u,v) \in E$ to a numeric value:
\begin{equation}
\emph{f}: E \mapsto \mathbb {R},\quad (u,v)\mapsto f(u,v)
\end{equation}

Generally speaking, weight functions play the role to map an unweighted graph into a weighted one. Given a weight function set consisting of \emph{n} weight functions, $F=\{f_{1},...,f_{n}\}$, a generic unweighted graph G could be mapped to a weighted graph set $D=\{G_{1},...G_{n}\}$, where $G_{i}=(V,E_{i})$.

\subsection{DEFINITION 2 (Partition)}

Given an undirected, weighted graph $\mathcal{G}=(V,E)$, $P=\{p_{1},...,p_{C}\}$ is a partition of the edges into \textbf{\emph{C}} communities. Each node $v \in V$ belongs to at least one communities, even an isolated node would itself constitute a community of single member. Communities as the result could be partitioned again recursively thus grouped into a hierarchical structure.

\subsection{DEFINITION 3 (Partition Support Matrix)}

Given a weighted graph set $D=\{G_{1},...G_{n}\}$, where $G_{i}=(V,E_{i})$, each weighted graph $G_{i}$ corresponds to a partition $P_{i}$. The partition support matrix of a node \emph{v} in partition $P_{i}$, written $PSM_{i}(v)$, is of column $\left\vert \emph{V} \right\vert$ and of row \emph{k} where $\left\vert \emph{V} \right\vert$ is the node number of graphs and k is the number of communities in partition $P_{i}$ that contains node \emph{v}. The element in \emph{i-th} row and \emph{j-th} column of $PSM_{i}(v)$ denotes whether \emph{j-th} node exists in \emph{i-th} communities which contains node \emph{v}. The partition support matrix of a node \emph{v} for all partitions, written $PSM(v)$, is defined as:

\begin{equation}
PSM(v)=
\left(\begin{array}{c}
PSM_{1}(v) \\
PSM_{2}(v) \\
...        \\
PSM_{n}(v)
\end{array}\right)
\end{equation}

\begin{figure}
\centering
\includegraphics[width=0.9\linewidth]{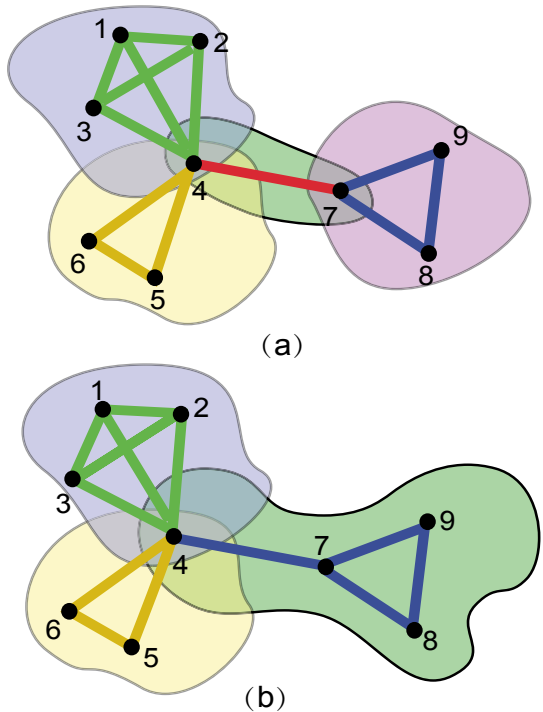}
\caption{Two different partition on the same set of variables}
\label{fig:psm}
\end{figure}

For example, the partition support matrix of the $node_{7}$ for partition $P_{1}$ in Figure \ref{fig:psm}a is $PSM_{1}(node_{7})=[[0,0,0,1,0,0,1,0,0]^T, [0,0,0,0,0,0,1,1,1]^T]^T$, while for partition $P_{2}$ in Figure \ref{fig:psm}b is $PSM_{2}(node_{7})=[0,0,0,1,0,0,1,1,1]$. Then the overall partition support matrix of $node_{7}$ is

\begin{equation}
PSM(node_{7})=
\left(\begin{array}{c}
0,0,0,1,0,0,1,0,0 \\
0,0,0,0,0,0,1,1,1 \\
0,0,0,1,0,0,1,1,1
\end{array}\right)
\nonumber
\end{equation}

 As one can see, $node_{7}$ tends to be grouped together with $node_{4}$, $node_{8}$ and $node_{9}$.

\subsection{DEFINITION 4 (Second-Order Network)}

Given a weighted graph set $D=\{G_{1},...G_{n}\}$, where $G_{i}=(V,E_{i})$, and its corresponding partition set $P=\{P_{1},...P_{n}\}$, the second-order network in an undirected,weighted graph $S=(V,E_{S})$ where an edge $e=(u,v) \in E_{S}$ if its edge weight exceeds threshold. The edge weight is valued by co-occurrence probability in partition support matrix between \emph{u} and \emph{v}.

\subsection{DEFINITION 5 (Second-Order Partition)}

A second-order partition is the partition $P_{sec}$ of a second-order network.

\section{\emph{LSBN} system: Large-Scale Bayesian Network Learning}

The \emph{LSBN} system provides a novel framework for Bayesian structure learning using model averaging in large-scale networks. The \emph{LSBN} system proposes a divide-and-conquer strategy to segment the originally intractable Bayesian structure learning tasks into multiple tractable sub-tasks with a much smaller scale. The workflow of \emph{LSBN} system is as follows:\\
(1) Variable Partition. (Section \ref{partition})  \\
(2) Sampling and Learning. (Section \ref{learn}) \\
(3) Mergence. (Section \ref{merge}) \\

\subsection{Variable Partition}
\label{partition}

The \emph{LSBN} system is expected to perform well even the number of variables increases. Variable partition serves as a crucial step by drastically reducing the complexity and run-time of learning in our next stage. Partition incorporates overlapping, that is, each node may belongs to more than one community. Ideally, a perfect partition should possess three properties: selectiveness, so that nodes within the communities have much higher probability to possess correct edges among each other than outside of the communities; high coverage, so that all correct edges are embodied within at least one community, in other words, partition doesn't break any correct edges; and fine granularity, so that each community is small enough to be applicable for Bayesian structure learning algorithms.

\begin{algorithm}
\caption{ROPART}
\label{alg:partition}
\begin{algorithmic}[1]
\State build an undirected, unweighted complete graph $\mathcal{G}$ given all variables;
\State for each weight function $f_{i}$ from a predefined weight function set $F=\{f_{1},...,f_{n}\}$, map $\mathcal{G}$ to a weighted graph $\mathcal{G}_{i}$ respectively;
\State for each weighted graph $\mathcal{G}_{i}$,keep those edges of which the weight exceed some truncate threshold $\mathcal {T}_{trunc}$, after that, generate partition $P_{i}$ respectively;
\State for each variable \emph{v}, construct its corresponding partition support matrix $PSM(\emph{v})$;
\State construct second-order network and generate second-order partition;

\end{algorithmic}
\end{algorithm}

\begin{figure*}[htb]
\centering
\includegraphics[width=1.0\textwidth]{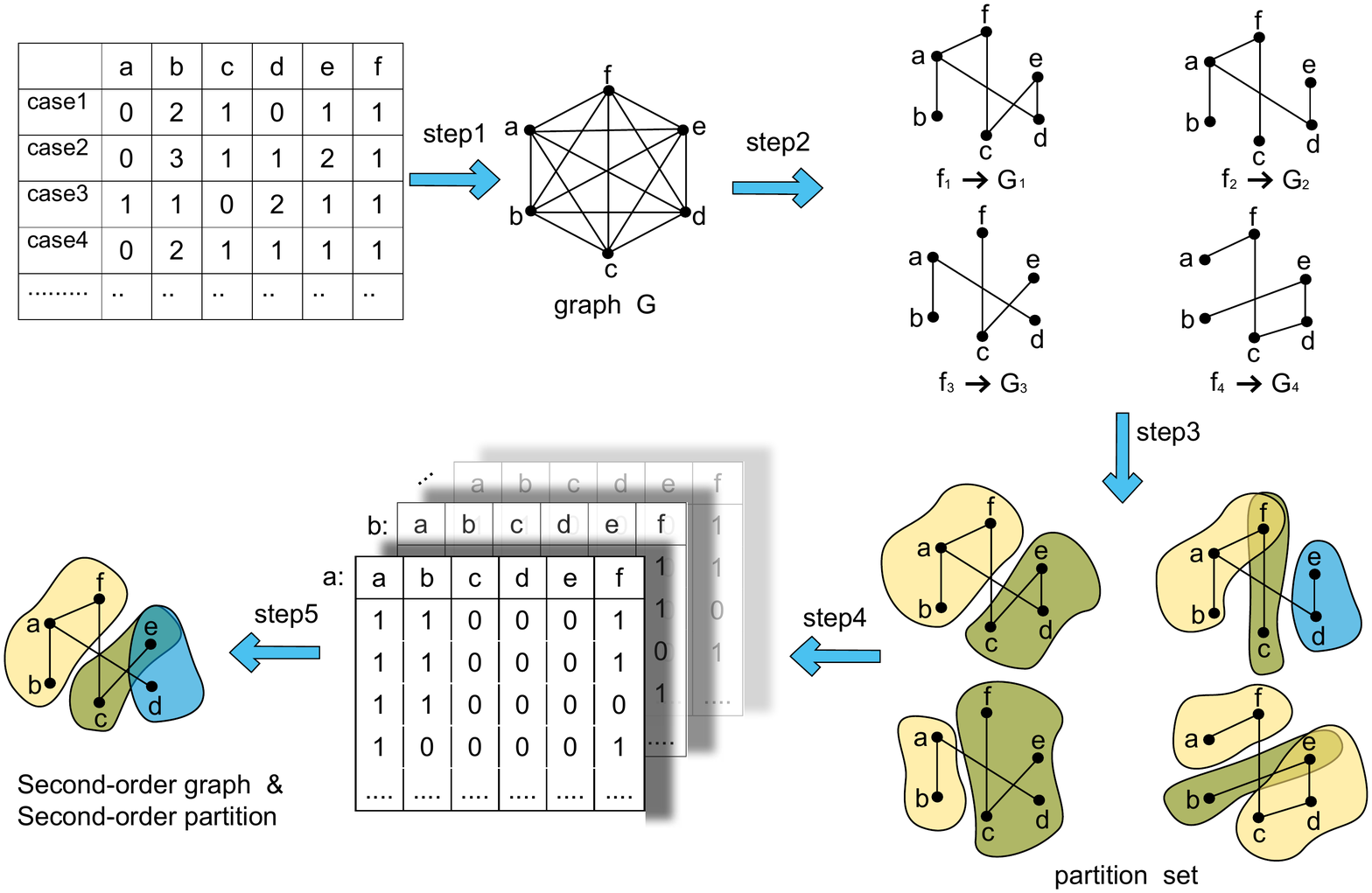}
\caption{ROPART: a robust partition strategy combining multiple different weight functions}
\label{fig:ropart}
\end{figure*}

The initial input of the \emph{LSBN} system requires discrete-state data. If the data is continuous, discretization is necessary before any further steps. We develop a novel partition algorithm, called 'ROPART', to construct robust partition of all variables. ROPART consists of five steps, as outlined in Algorithm \ref{alg:partition} and illustrated in Figure \ref{fig:ropart}.\\

In step 1, We start with a fully-connected, undirected, unweighted graph including all variables. \\

In step 2, We employ a predefined weight function set $F=\{f_{1},...,f_{n}\}$ to translate original unweighted graph into a set of various weighted graphs $D=\{\mathcal{G}_{1},...\mathcal{G}_{n}\}$, where $f_{i}$ corresponds to $\mathcal{G}_{i}$. The predefined weight functions are shown as follows:\\

$\bullet$ Mutual Information
\begin{equation}
\begin{aligned}
MI(X,Y)=\sum_{x}\sum_{y}P(x,y)log\frac{P(x,y)}{P(x)P(y)}
\nonumber
\end{aligned}
\end{equation}

$\bullet$ Mutual Information normalized by the sum of entropies \cite{WittenF05}
\begin{equation}
\begin{aligned}
MI_{plus}(X,Y)=\frac{2MI(X,Y)}{H(X)+H(Y)}
\nonumber
\end{aligned}
\end{equation}

$\bullet$ Mutual Information normalized by the square root product of entropies\cite{DBLP:conf/aaai/StrehlG02}
\begin{equation}
\begin{aligned}
MI_{sqrt}(X,Y)=\frac{MI(X,Y)}{\sqrt{H(X)}\sqrt{H(Y)}}
\nonumber
\end{aligned}
\end{equation}

$\bullet$ Mutual Information normalized by PageRank weight
\begin{equation}
\begin{aligned}
MI_{pr}(X,Y)=\frac{MI(X,Y)}{\sqrt{PR(X)}\sqrt{PR(Y)}}
\nonumber
\end{aligned}
\end{equation}

where \emph{PR(X)} and \emph{PR(Y)} are PageRank values of node \emph{X} and \emph{Y} respectively.

$\bullet$ Mutual Information by Standard Normalization
\begin{equation}
\begin{aligned}
MI_{sn}(X,Y)=\frac{MI(X,Y)-\mu _{MI}}{\sigma _{MI}}
\nonumber
\end{aligned}
\end{equation}

where $\mu _{MI}$ and $\sigma _{MI}$ denotes the mean value and standard deviation of all edge weights, which are valued by mutual information.

$\bullet$ Pearson Correlation Coefficient
\begin{equation}
\begin{aligned}
\rho_{X,Y}=\frac{E[(X-\mu_{X})(Y-\mu_{Y})]}{\sigma_{X} \sigma_{Y} }
\nonumber
\end{aligned}
\end{equation}

where $\mu_{X}$ and $\mu_{Y}$ denote the mean value of variable $X$ and $Y$ respectively, and $\sigma_{X}$ and $\sigma_{Y}$ indicate the standard deviation of $X$ and $Y$ correspondingly.

Each weighted graph $\mathcal{G}_{i}$ derived by weight function $f_{i}$ is pruned by removing edges whose weight is lower than some truncate threshold $\mathcal {T}_{trunc}$.\\

In step 3, we partition each weighted graph $\mathcal{G}_{i}$ after pruning. For the sake of convenience in Mergence(Section \ref{merge}), we prefer communities to be organized hierarchically. Meanwhile, for the sake of high coverage, we wish communities to maintain pervasive overlaps. Here we use LinkComm\cite{YY_LC_nature2010} for partition algorithm, which is introduced in Section \ref{relate}.

In step 4, We construct partition support matrix $PSM(\emph{v})$ for each variable \emph{v}, based on partition result set $P=\{P_{1},...P_{n}\}$ generated from step 3, where $P_{i}$ corresponds to weighted graph $\mathcal{G}_{i}$. \\

In step 5, We build second-order network based on partition support matrix set $PSM=\{ PSM(\emph{v}_{1}),...PSM(\emph{v}_{\left | V \right |})\}$ and generate second-order partition $P_{sec}$. The resulting second-order partition will satisfy the criteria for a good partition: (1) the communities are of high cohesion and low coupling; and (2) the partition is robust to the selection of weight function.\\

\subsection{Sampling and Learning}
\label{learn}

\begin{figure}[ht]
    \centering
    \subfigure[]{
        \includegraphics[width=.45\linewidth]{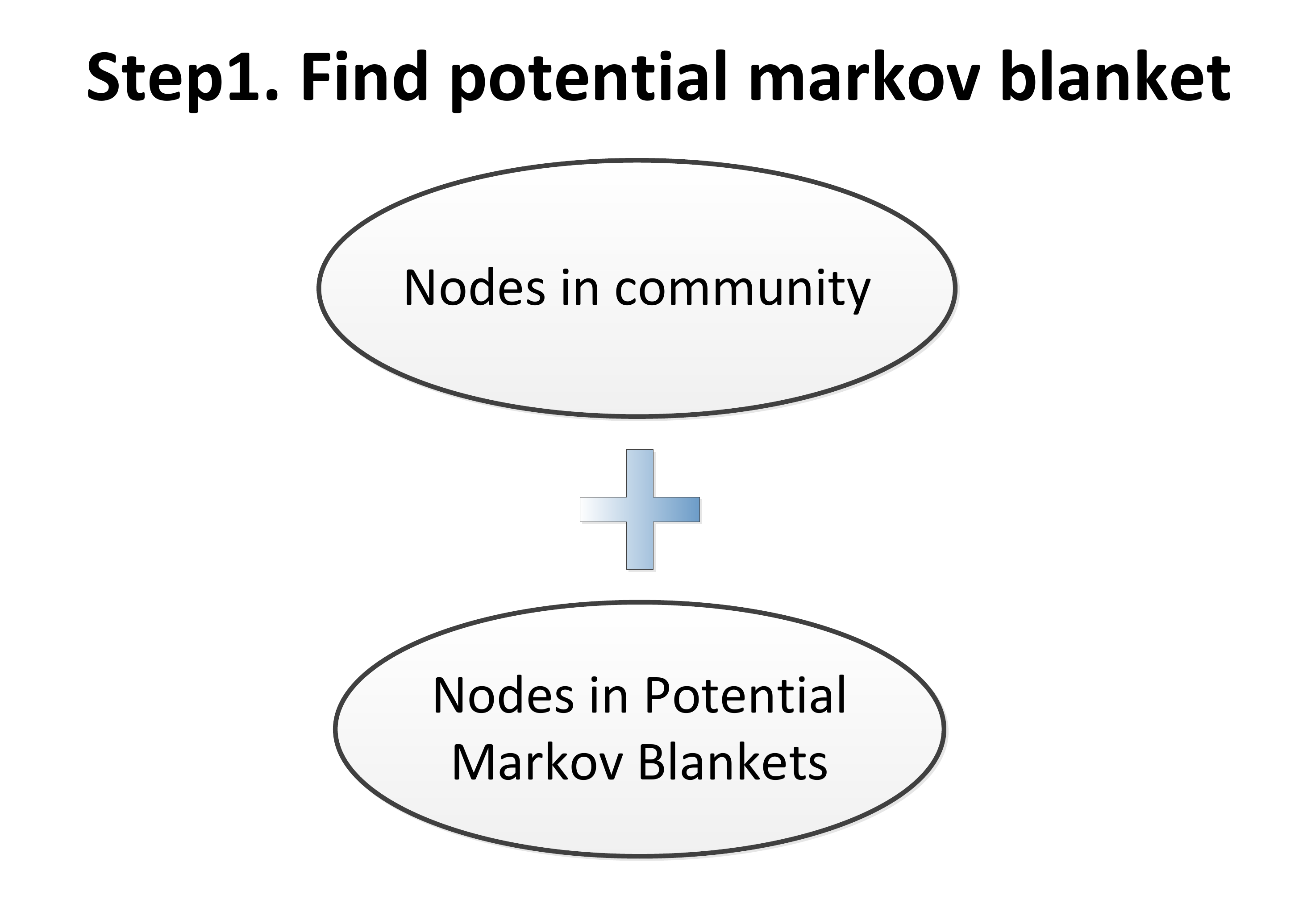}
        \label{fig:learn1}
        }
    \subfigure[]{
        \includegraphics[width=.45\linewidth]{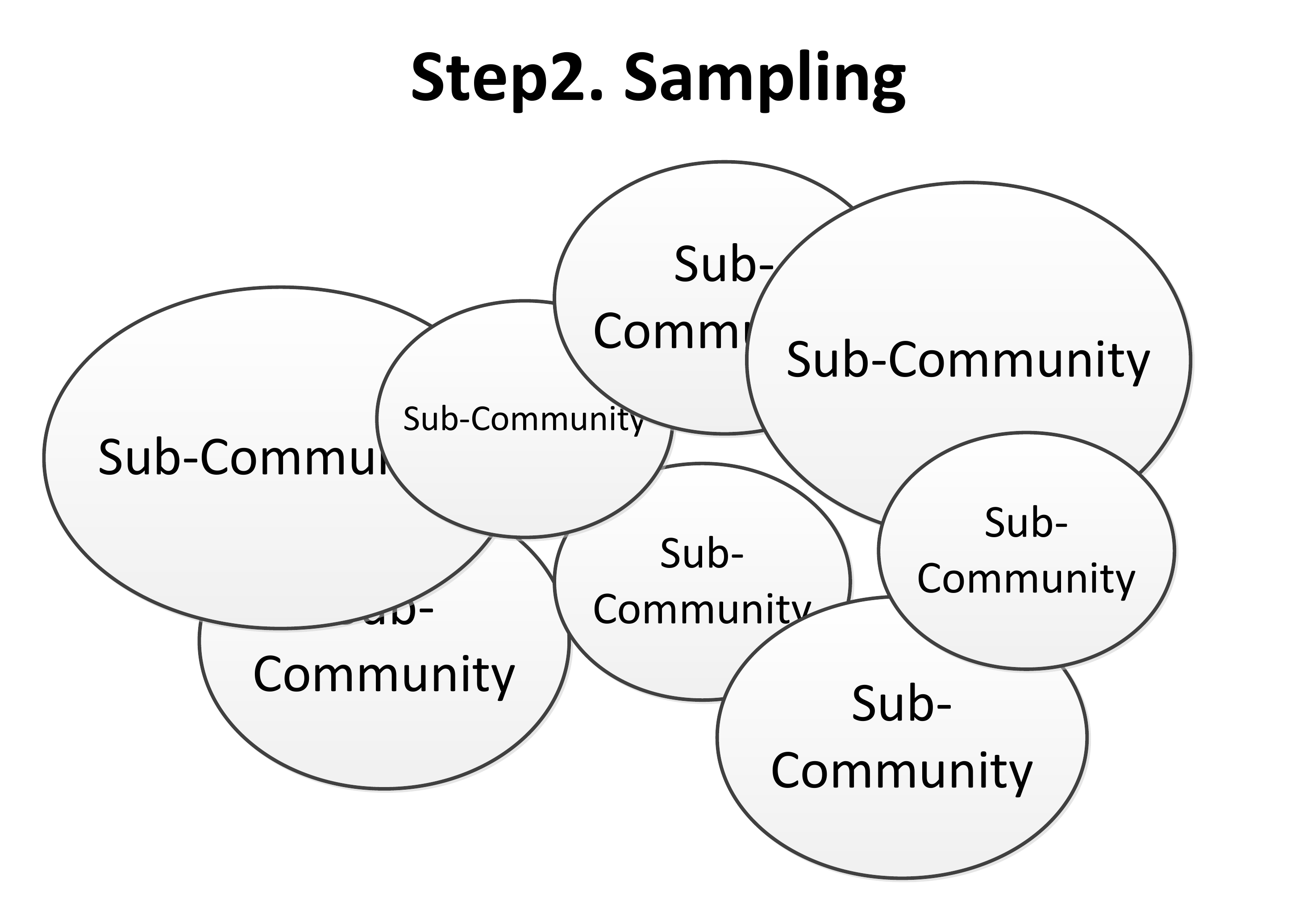}
         \label{fig:learn2}
        }
    \subfigure[]{
        \includegraphics[width=.45\linewidth]{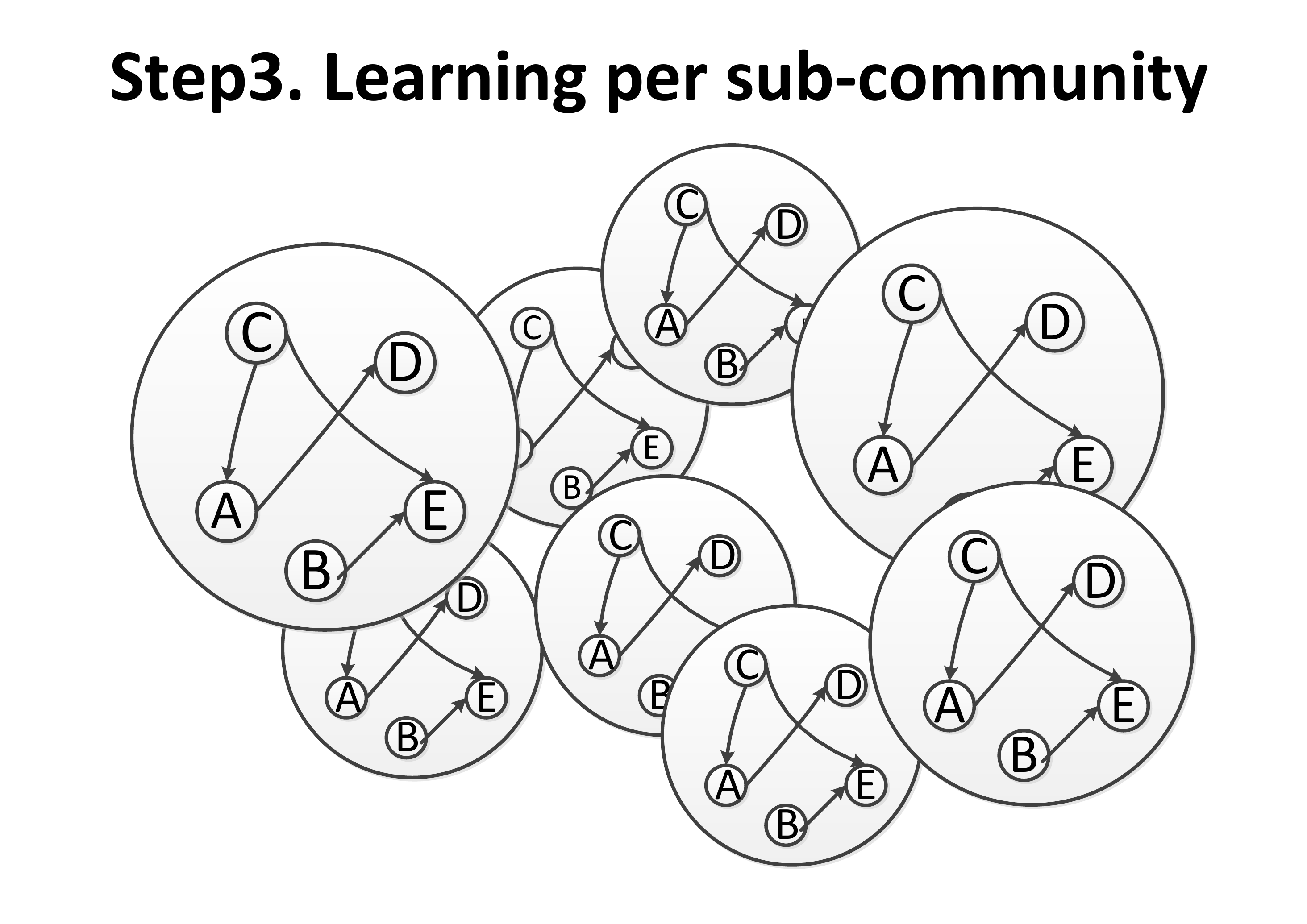}
         \label{fig:learn3}
        }
    \subfigure[]{
        \includegraphics[width=.45\linewidth]{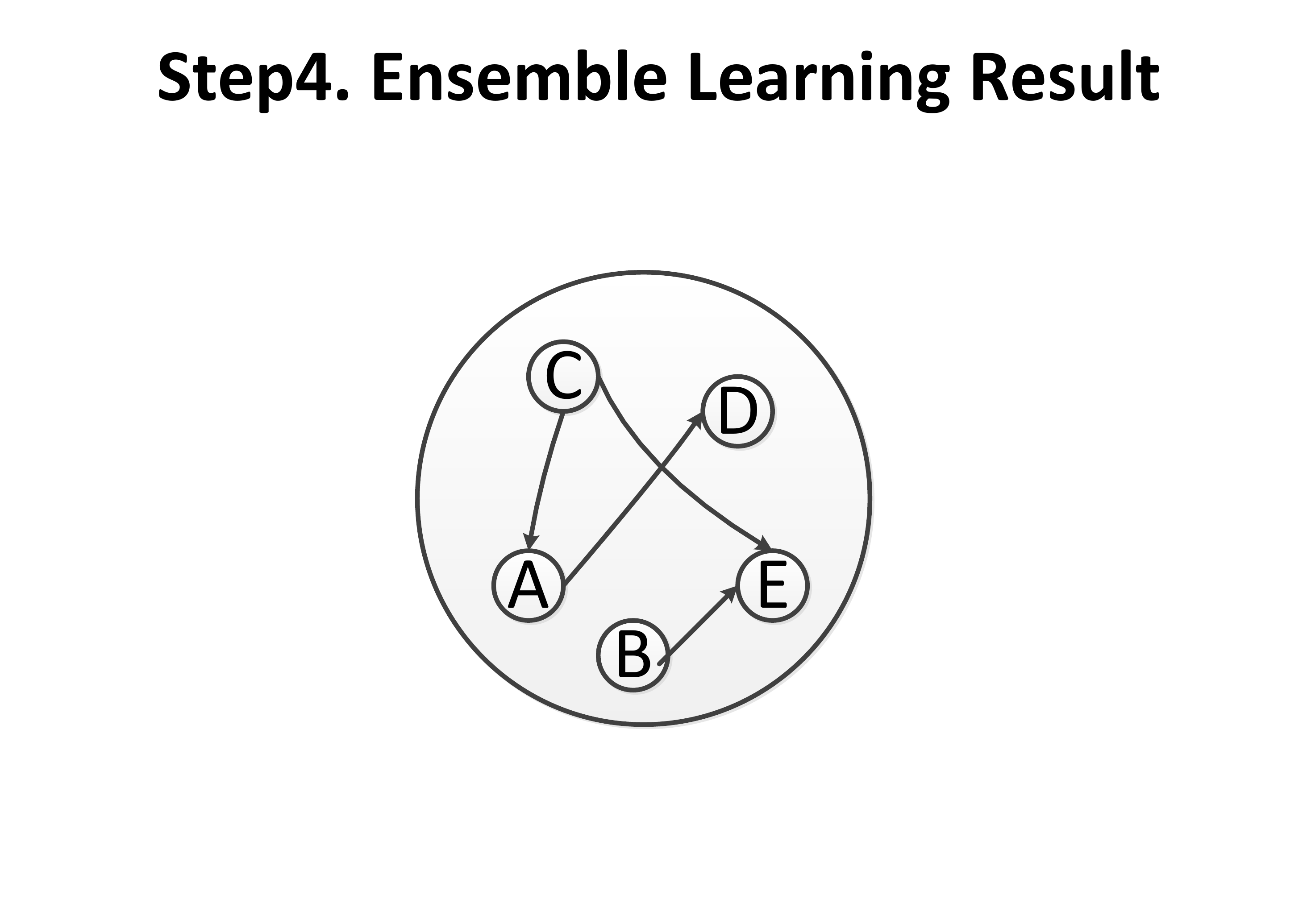}
         \label{fig:learn4}
        }
    \caption{LocalLearn: a sampling strategy for bayesian structure learning per community}
    \label{fig:learn}
\end{figure}

The \emph{LSBN} system performs Bayesian structure learning per community, however, to learn the structure correctly poses many challenges. For example, what if an outlier variable is erroneously mixed into a community, how to pinpoint such variables, and what's most important, how to eliminate the learning bias caused by irrelevant variables in the same community? What's more, what if correct edges suffer disconnection during partition, is there any mechanism to retrieve those missing edges? We first attempt to find out candidate Markov Blanket given nodes within community, after that, we consider a sampling methodology to address mentioned challenges. The detailed learning algorithm consists of four steps, as outlined in Algorithm \ref{alg:learn} and illustrated in Figure \ref{fig:learn}.\\

\begin{algorithm}
\caption{LocalLearn}
\label{alg:learn}
\begin{algorithmic}[1]
\For{each community $\mathcal {C} \in P_{sec}$}
    \State identify candidate Markov Blanket $MB(\mathcal {C})$
    \State sample \emph{k} sub-communities based on $\mathcal {C}$ and $MB(\mathcal {C})$, $SC=\{sub_{1}(\mathcal {C}), ..., sub_{k}(\mathcal {C})\}$
    \For{each sub-community $sc_{i} \in SC$}
        \State learn Bayesian structure
    \EndFor
    \State ensemble structures of all sub-communities and resolve conflicts
\EndFor
\end{algorithmic}
\end{algorithm}

In Step 1, we try to find out the Markov Blanket $MB(\mathcal {C})$ for community $\mathcal {C}$. Conditioned on Markov Blanket, no other nodes outside the community $\mathcal {C}$ could influence nodes within the community. To be more precise, $\forall X \in C,X\perp Y| MB(X)$ for all $Y \notin \left \{ X \cup MB(X) \right \}$. Therefore knowledge of Markov Blanket $MB(\mathcal {C})$ is enough to infer intra-community structure, thus makes the learning problem localized and easier to be solved.

According to definition, the Markov Blanket of variable \emph{X} is composed of all parents of \emph{X}, all children of \emph{X} and all parents of \emph{X}'s children. In other words, Markov Blanket of \emph{X} should be closer to \emph{X} topologically than any other variables in networks.

Since the Markov Blanket of variable \emph{X} is composed of all parents of \emph{X}, all children of \emph{X} and all parents of \emph{X}'s children, in other words, Markov Blanket of \emph{X} should be closer to \emph{X} topologically than any other variables in structure. Besides, topological closeness is related to significance in edge weights, thus edge weights play an important role in identifying Markov Blanket. In addition, conditional independence (CI) tests and measurement of association among variables are required for further justification. The identification of Markov Blanket is divided into two steps: \\

(1) for variable \emph{X}, we achieve its Markov Blanket candidates by looking for adjacent nodes whose edge weight exceeds the average. To be precise, $MB_{cand}(X)=\{ Y | w_{XY} \geq \overline{w}_{X} \}$, where $w_{XY}$ is the weight of edge $e_{XY}$ and $\overline{w}_{X}$ is the average weight of edges connecting \emph{X}.\\

(2) Given $MB_{cand}(X)$, we employ a scalable algorithm IAMB\cite{iamb} for further justification. IAMB\cite{iamb}(Incremental Association Markov Blanket) consists of a forward phase and a backward phase. In the forward phase, IAMB applies a heuristic approach to find an estimated Markov Blanket set \emph{CMB}. \emph{CMB} starts with an empty set, then iteratively added variables \emph{Y} from $MB_{cand}(X)$ which maximizes a heuristic function $f(Y;X | CMB)$. The heuristic function $f(Y;X | CMB)$ denotes the mutual information between candidate Markov Blanket node \emph{Y} and target node \emph{X} given \emph{CMB}, which is informative and efficient. In the backward phase, conditional independence tests are used within \emph{CMB}, and invalid variables are removed from \emph{CMB} one by one if it is independent of \emph{X} given the remaining \emph{CMB}.\\

In Step 2, we combine each community $\mathcal {C}$ with its Markov Blanket $MB(\mathcal {C})$ into a expanded community ${\mathcal {C}}'$. The size of ${\mathcal {C}}'$ would be much smaller than expectation, for intra-community variables are highly correlated due to partition. Chances are high that members of Markov Blanket is embodied in community $\mathcal {C}$ as well.

Given expanded community ${\mathcal {C}}'$, we conduct sampling for two reasons: (1) the size of ${\mathcal {C}}'$ may still be too large for practical Bayesian structure learning; (2) bootstrap contributes to more consistent and robust results. Our sampling algorithm borrows the idea of Random Node Neighbor (RNN)\cite{leskovec2006sampling} with some modifications.

In our sampling algorithm, we build up an inner markov graph $\mathcal{G}_{IM}(\mathcal {C})$ based on $\mathcal {C}$, to be precise, $\mathcal{G}_{IM}(\mathcal {C})=(\mathcal {C},E)$ where $e_{XY}=1$ if $X \in MB(Y)$ or $Y \in MB(X)$, and 0 otherwise. Then we uniformly pick an unvisited node from inner markov graph $\mathcal{G}_{IM}(\mathcal {C})$ as starting node at random together with its neighbors, denoted by $\mathcal {S}$. The final sub-community for Bayesian structure learning is $\mathcal {S} \cup MB(\mathcal {S})$, if the size of ultimate sub-community is still too large to learn, we keep removing neighbors within $\mathcal {S}$ until acceptable size.\\

In Step 3, for each sub-community, LocalLearn use Bayesian model averaging to learn the structure of nodes within the sub-community. Bayesian model averaging is different from other structure learning algorithms, such as constrain-based approach and score-and-search approach, for it is interested in the confidence of some structure-related features rather than structure topology per se. In this case, feature $f(\mathcal{G})=1$ denotes whether there exists an edge from node \emph{i} to node \emph{j} and $f(\mathcal{G})=0$ otherwise. Usually posterior expectation of feature $f(\mathcal{G})$ is estimated by:\\
\begin{equation}
\begin{aligned}
P(f|D)=\sum_{G}f(G)p(G|D)
\end{aligned}
\label{eq:origBNAvg}
\end{equation}

$f(\mathcal{G})$ is classified to be 1 if $E(f|D)$ exceed certain threshold $\mathcal {T}_{avg}$ and 0 otherwise. However, traversing all candidate DAGs according to Eq.(\ref{eq:origBNAvg}) is usually intractable for there are overall $O(n!2^{\binom{n}{2}})$ DAGs given \emph{n} nodes. One solution is to average over structures which conform to some predetermined node order $\prec$ \cite{DBLP:journals/ml/CooperH92}. For example, if $X_{i} \in Pa_{G}(X_{j})$ then $i \prec j$. Now the posterior expectation of feature $f(\mathcal{G})$ given a predetermined order $\prec$ could be transformed into\cite{Friedman03}: \\
 \begin{equation}
\begin{aligned}
P(f|D)=\sum_{\prec }P(\prec|D)P(f|D, \prec)
\end{aligned}
\label{eq:orderBNAvg}
\end{equation}

Usually the order $\prec$ is unobtainable due to little prior in current domain, so there are $n!$ possible orders in all to be taken into consideration given \emph{n} nodes, which remains intractable. So we use MCMC (Markov Chain Monte Carlo) techniques to sample orders, and sample DAGs consistent with that order\cite{Friedman03}\cite{Eaton07_uai}. Assuming a uniform prior over orders, a Markov Chain $\mathcal {M}$ is constructed with state space consisting of all $n!$ possible orders. This Markov Chain $\mathcal {M}$ is simulated by conforming to stationary distribution $P(\prec|D)$, and a sequence of sampled order $\prec_{1},...,\prec_{T}$ is obtained. Now the expectation of feature f could be approximately estimated as\cite{Friedman03}: \\
 \begin{equation}
\begin{aligned}
P(f|D)\approx \frac{1}{T}\sum_{t=1}^{T}P(f|D, \prec_{t})
\end{aligned}
\label{eq:orderMCMCBNAvg}
\end{equation}

After learning sub-community structures, we integrate them into a uniform intra-community structure and resolve conflicts. We investigate the characteristics of error edges and find two major types of error:

$\bullet$ additional edges due to indirect interactions. This type of error is introduced by ARACNE\cite{DBLP:journals/bmcbi/MargolinNBWSFC06}. ARACNE\cite{DBLP:journals/bmcbi/MargolinNBWSFC06} tries to eliminate indirect edges by using Data Processing Inequality in all triplets of variables. However, ARACNE fails to eliminate all indirect edges and keep all valid edges at the same time for two reasons: (1) it is hardly to find fixed threshold to distinguish indirect interactions from direct ones; (2) In many cases, weakest edges in triplets do not necessarily indicate indirect interaction.

$\bullet$ missing edges due to weak edge weight. This type of error originates from the adjacent nodes, maybe some nodes have relatively small PageRank value, in other words, they are periphery nodes; Or maybe some nodes are hubs, so the mutual information with its neighbors, that is, the edge weight, is relatively small.

To tackle these two major types of error, we first collect all candidate triplets of nodes. A candidate triplet contains three nodes, mutually-connected by the edge whose weight exceeds certain threshold value. Chances are high that these candidate triplets contain indirect interactions\cite{DBLP:journals/bmcbi/MargolinNBWSFC06}. We build up a undirected, unweighted graph based on edges from candidate triplets. Then we cluster this graph into sparsely-connected dense subgraphs. Ideally, on one hand, indirect interactions are clustered with direct interactions. Since direct interactions have more significant edge weights, indirect interactions would be eliminated through re-learning. On the other hand, weak correct edges are expected to be grouped with weaker wrong edges. As a result, they survive through re-learning due to relative significance. Here we still employ LinkComm\cite{YY_LC_nature2010} for graph clustering. After clustering the candidate triplet graph, we re-learn the structure for each cluster using the same method in Step 3.

\subsection{Mergence}
\label{merge}

\begin{algorithm}
\caption{Mergence}
\label{alg:merge}
\begin{algorithmic}[1]
\State Create a leaf node for every community and add it to the Community Pool;
\While{there is more than one node in community pool}
    \State Remove nodes $\mathcal {C}_{i}$ and $\mathcal {C}_{j}$ of maximum Jaccard similarity coefficient $\mathcal {J}(\mathcal {C}_{i}, \mathcal {C}_{j})$ from the Community Pool;
    \State Merge structures of two communities $\mathcal {C}_{i}$ and $\mathcal {C}_{j}$;
    \State Create a new node $\mathcal {C}_{new}=\mathcal {C}_{i}\bigcup \mathcal {C}_{j}$ denoting the mergence of two communities $\mathcal {C}_{i}$ and $\mathcal {C}_{j}$;
    \State Add the new node to the Community Pool;
\EndWhile
%\State The remaining node is the complete set of nodes, with its corresponding structure.
\end{algorithmic}
\end{algorithm}

The \emph{LSBN} system combines structures after learning individually from each community. The combination involves two concerns: (1) to find an efficient mergence order; (2) to resolve the conflicts during the mergence. Intuitively, the mergence strategy should proceed as a bottom-up approach. The \emph{LSBN} system would keep piecing together all intra-community structures into larger structures, block by block until a whole structure is achieved. We expect the structure to be increasingly accurate and wrong edges would be continuously eliminated during the mergence.

By borrowing the idea of Huffman's Algorithm\cite{huf52}, the \emph{LSBN} system tries to merge communities in a greedy strategy by constantly pick two communities with maximum Jaccard similarity coefficient\cite{jaccard}. Jaccard similarity coefficient\cite{jaccard} of two communities is proportional to their overlap and inversely proportional to their union size. We propose Mergence Algorithm to perform such greedy strategy in order to combine the structures of each community into whole better, and the detailed algorithm is outlined in Algorithm \ref{alg:merge}.

At first, Mergence Algorithm put all communities into a Community Pool, then repeatedly choose two communities $\mathcal {C}_{i}$ and $\mathcal {C}_{j}$ with largest Jaccard similarity coefficient $\mathcal {J}(\mathcal {C}_{i}, \mathcal {C}_{j})=\frac{\left | \mathcal {C}_{i}\bigcap \mathcal {C}_{j} \right |}{\left | \mathcal {C}_{i}\bigcup \mathcal {C}_{j}\right |}$. After two communities are selected, same approach described in step 3 of Section \ref{learn} is applied to resolve the conflicts by clustering triplets. Then Mergence Algorithm combine these two communities into a new hybrid community, and put it into the Community Pool for further mergence steps.

In each iteration, assuming there are \emph{k} communities left in the Community Pool, then it would take $\binom{k}{2}$ times of calculation to select the maximum value of Jaccard similarity coefficient. If there are \emph{n} communities initially, the total calculation sums up to be:  $\sum_{k=2}^{n}\binom{k}{2}=\frac{1}{2}[\sum_{k=2}^{n} k^{2}-\sum_{k=2}^{n}k]=\frac{n(n+1)(n-1)}{6}$. For the sake of computational efficiency, the values of Jaccard similarity coefficient could be calculated in advance. For each iteration, assuming there are \emph{k} communities left in the Community Pool, removing old values would take only $2(k-1)+1$ times and adding new values would take $k-2$ times of calculation. After computational optimization, the overall calculation shrinks to: $\binom{n}{2}+\sum_{k=2}^{n}[2(k-1)+1+(k-2)]=2n(n-1)$.

\section{Experiments}
\label{experiments}

We benchmark evaluation of \emph{LSBN} on five well-known datasets. We expect the structures learned by \emph{LSBN} to be close enough to those learned by other Bayesian structure learning algorithms. Closeness in results indicates that partition and local learning in \emph{LSBN} hardly cause any losses. In addition, since \emph{LSBN} is designed to work on Bayesian structure learning problem in large-scale network, we expect \emph{LSBN} to learn structures of which the sizes exceed the computational upper bound of traditional Bayesian model averaging approach.

\subsection{Datasets}
\label{exp:dataset}

$\bullet$ alarm \cite{alarm}. The alarm network consists of 37 random variables and 46 arcs, with average degree of network being 2.49, maximum in-degree being 4 and average Markov Blanket size being 3.51.\\

$\bullet$ insurance \cite{insurance}. The insurance network contains 27 random variables and 52 arcs, with average degree of network being 3.85, maximum in-degree being 3 and average Markov Blanket size being 5.19.\\

$\bullet$ win95pts \cite{win95pts}. The win95pts network includes 76 random variables and 112 arcs, with average degree of network being 2.95, maximum in-degree being 7 and average Markov Blanket size being 5.92.\\

$\bullet$ pigs \cite{pigs}. The pigs network includes 441 random variables and 592 arcs, with average degree of network being 2.68, maximum in-degree being 2 and average Markov Blanket size being 3.66.\\

$\bullet$ link \cite{link}. The link network embodies 724 random variables and 1125 arcs, with average degree of network being 3.11, maximum in-degree being 3 and average Markov Blanket size being 4.8.\\

From each benchmark network, we sampled 20000 instances as the observed data. Besides, the pre-defined weight function set includes: (1) mutual information, abbreviated as 'MI'; (2) mutual information normalized by the sum of entropies, abbreviated as '$MI_{plus}$'; (3) mutual information normalized by the square root product of entropies, abbreviated as '$MI_{sqrt}$'; (4) mutual information normalized by the PageRank values, abbreviated as '$MI_{pr}$'; (5) mutual information after standard normalization, abbreviated as '$MI_{sn}$'; (6) Pearson Coefficient in absolute value, abbreviated as 'Pearson'; (7) absolute Pearson Coefficient after standard normalization, abbreviated as '$Pearson_{sn}$'.

\subsection{Parameter Setting}
\label{exp:ParamSet}

For each weighted network generated by certain weight function, the truncate threshold $\mathcal {T}_{trunc}$ for pruning is chosen by the Elbow Method\cite{Thorndike53whobelongs}. Specifically, we first transform all edge weights into histogram, and each bin in the histogram denotes the frequency of edges weights falling in certain range. Then we look at the variance descent between each pair of adjacent bins. For example, the first bin will add much information (encompass a lot of variance), for the majority of the edges possess relatively very small weights. Yet at certain bin, the variance ratio slows down. And we choose that bin as the truncate threshold $\mathcal {T}_{trunc}$, also called 'elbow criterion'. The truncate thresholds we chosen are shown in Table \ref{tab:threshold}, the percentage denotes the ratio between remaining edges and all edges in complete graph. From the results in Table \ref{tab:threshold}, there is no significant difference in numbers of remaining edges.

We measure the average shortest path (Table \ref{tab:shortestPath}) and diameter (Table \ref{tab:maximumPath}) for each partition, in comparison of weighted network partitions and second-order partition. There are several noticeable phenomenon in these results. First, there is no dominating weight function, for the partition result of its weighted network performs excellent in some datasets but poor in others. For example, mutual information proves to be the best weight function in alarm according to average shortest path, yet among the worst ones in win95pts. Second, as expected, second-order partition achieves more stable results. The results always belong to the best ones and never oscillate drastically. The average ranking for second-order partition in average shortest path is 2.4 (Table \ref{tab:shortestPath}), and the average ranking in diameter is 2.

The partition results are depicted in Table \ref{tab:partitionDist}, and the partition size distribution reveals that the size of the majority of communities ranges less or equal than 25 (100\% in alarm, 100\% in insurance, 100\% in win95pts, 92.827\% in pigs and 95.181\% in link). The average community size is 10.286 in alarm, 9.8 in insurance, 8.345 in win95pts, 9.527 in pigs and 8.904 in link.

\begin{table*}[htb]
\caption{\label{tab:threshold}Truncate threshold for each weight function on each dataset}
\begin{center}
\begin{tabular}{|l||c|c||c|c||c|c||c|c||c|c|}
\hline
 &\multicolumn{2}{c||}{insurance}&\multicolumn{2}{c||}{alarm}&\multicolumn{2}{c||}{win95pts}&\multicolumn{2}{c||}{pigs}&\multicolumn{2}{c|}{link}\\
\cline{2-11}
 & $\mathcal {T}_{trunc}$ & percent & $\mathcal {T}_{trunc}$ & percent & $\mathcal {T}_{trunc}$ & percent & $\mathcal {T}_{trunc}$ & percent & $\mathcal {T}_{trunc}$ & percent\\
\hline
MI             & 0.08 & 0.2308 & 0.01 & 0.2267 & 0.007 & 0.1659 & 0.005 & 0.0841 & 0.035 & 0.0566 \\
$MI_{plus}$    & 0.2  & 0.2792 & 0.03 & 0.2312 & 0.05  & 0.1635 & 0.05  & 0.0958 & 0.4   & 0.0435 \\
$MI_{sqrt}$    & 0.5  & 0.3646 & 0.5  & 0.2132 & 0.4   & 0.1618 & 2     & 0.0726 & 2.5   & 0.0493 \\
$MI_{pr}$      & 0.05 & 0.3276 & 0.02 & 0.2267 & 0.013 & 0.1687 & 0.005 & 0.0762 & 0.5   & 0.0510 \\
$MI_{sn}$      & 0    & 0.2707 & 0    & 0.2072 & 0.05  & 0.1368 & 0.8   & 0.0861 & 0.6   & 0.0441 \\
Pearson        & 0.1  & 0.4017 & 0.4  & 0.2747 & 0.02  & 0.1894 & 0.026 & 0.1516 & 0.075 & 0.0585 \\
$Pearson_{sn}$ & 0    & 0.3789 & 0    & 0.2087 & 0.75  & 0.1175 & 0     & 0.0731 & 1.0   & 0.0490 \\
\hline
\end{tabular}
\end{center}
\end{table*}

\begin{table*}[htb]
\caption{\label{tab:shortestPath}Comparison of averaging shortest path for various weight functions}
\begin{center}
\begin{tabular}{|l|cccccc|}
\hline
 & insurance & alarm & win95pts & pigs & link & $ranking_{avg}$ \\
\hline
MI            & 1.5333(3) & 3.8162(1) & 2.9799(6) & 5.2616(4) & 2.6427(4) & 3.6 \\
$MI_{plus}$   & 1.55(4)   & 4.1567(8) & 3.1130(8) & 5.4789(5) & 2.4552(1) & 5.2 \\
$MI_{sqrt}$   & 1.9456(8) & 3.8792(3) & 3.0156(7) & 5.6248(7) & 2.4622(2) & 5.4 \\
$MI_{pr}$     & 1.7983(6) & 3.8567(2) & 2.5521(1) & 5.1297(3) & 3.7409(6) & 3.6 \\
$MI_{sn}$     & 1.4927(2) & 3.9682(7) & 2.8828(5) & 5.5749(6) & 4.4982(8) & 5.6 \\
Pearson       & 1.7357(5) & 3.8991(5) & 2.7558(2) & 4.4463(2) & 2.6801(5) & 3.8 \\
$Pearson_{sn}$& 1.8430(7) & 3.9178(6) & 2.8380(4) & 5.6633(8) & 4.0882(7) & 6.4\\
second-order& 1.36(1)   & 3.8875(4) & 2.8240(3) & 4.28198(1)& 2.5624(3) & \textbf{2.4}\\
\hline
\end{tabular}

Note: The number within the parentheses denotes the ascending ranking of current weight function in certain dataset.

\end{center}
\end{table*}

\begin{table*}[htb]
\caption{\label{tab:maximumPath}Comparison of averaging Diameters for various weight functions}
\begin{center}
\begin{tabular}{|l|cccccc|}
\hline
 & insurance & alarm & win95pts & pigs & link & $ranking_{avg}$ \\
\hline
MI      & 3.0(3)        & 6.1667(2) & 5.1429(6) & 10.6875(4) & 5.0026(8) & 4.6 \\
$MI_{plus}$ & 3.0(3)        & 6.8(7)    & 5.4(8)    & 11.2247(5) & 4.3666(4) & 5.4 \\
$MI_{sqrt}$ & 4.0(8)        & 6.8333(8) & 5.1667(7) & 11.672(8)  & 4.3480(3) & 6.8 \\
$MI_{pr}$   & 3.6(6)        & 6.2(3)    & 4.05(1)   & 9.4707(3)  & 4.6027(6) & 3.8 \\
$MI_{sn}$   & 2.7143(2)     & 6.7143(6) & 4.6857(4) & 11.4805(7) & 4.4982(5) & 4.8 \\
Pearson     & 3.33(5)       & 6.1(1)    & 4.5556(2) & 8.3585(2)  & 4.9513(7) & 3.4 \\
$Pearson_{sn}$ & 3.778(7)& 6.625(5)  & 4.886(5)  & 11.3907(6) & 4.0882(2) & 5 \\
second-order& 2.423(1)& 6.4211(4) & 4.619(3)  & 7.7565(1)  & 3.7409(1) & \textbf{2} \\
\hline
\end{tabular}

Note: The number within the parentheses denotes the ascending ranking of current weight function in certain dataset.

\end{center}
\end{table*}

\begin{table*}[htb]
\caption{\label{tab:partitionDist}Partition Size Distribution for all Datasets}
\begin{center}
\begin{tabular}{|c|c|c|c|c|c|c|c|c|c|c|c|}
\hline
 & 1-5 & 6-10 & 11-15 & 16-20 & 21-25 & 26-30 & 31-35 & 36-40 & 41-45 & 46-50 & $>$50 \\
\hline
alarm      & 3 & 2 & 1 & 2 & 0 & 0 & 0 & 0 & 0 & 0 & 0  \\
insurance  & 1 & 3 & 6 & 0 & 0 & 0 & 0 & 0 & 0 & 0 & 0  \\
win95pts   & 19& 19& 13& 4 & 0 & 0 & 0 & 0 & 0 & 0 & 0  \\
pigs       &102& 76& 24& 11& 7 & 6 & 6 & 1 & 1 & 1 & 2  \\
link       & 81& 47& 16& 5 & 9 & 3 & 0 & 2 & 3 & 0 & 0  \\
\hline
\end{tabular}
\end{center}
\end{table*}

\subsection{Experimental Design}
\label{exp:expDesign}

Our evaluation benchmarks \emph{LSBN} four state-of-art large-scale network structure learning algorithms, namely ARACNE\cite{DBLP:journals/bmcbi/MargolinNBWSFC06}, PC\cite{Spirtes2000}, Greedy Search\cite{Brown05acomparison} and Max-Min Hill Climbing (MMHC)\cite{Tsamardinos06themax-min}. Among these common algorithms, ARACNE\cite{DBLP:journals/bmcbi/MargolinNBWSFC06} is a very popular information-theoretic algorithm with extreme simplicity and low computational cost; the PC algorithm\cite{Spirtes2000} is considered as the most popular constrain-based algorithm; Greedy Search\cite{Brown05acomparison} are very widely-used score-and-search approaches; and MMHC\cite{Tsamardinos06themax-min} serves as a hybrid method which proves to be superior to other algorithms in most cases.

We compare \emph{LSBN} to these four state-of-art algorithms with respect to its correctness in structures. In implementation of PC, Greedy Search and MMHC, we were aided by Causal Explorer Toolkit\cite{CasualExplorer} (http://www.dsl-lab.org/causal\_explorer/) and structural results in the format of directed edges are evaluated by ourselves. The parameters used in Causal Explorer Toolkit for each algorithm are just set up as default, for example, threshold on statistical test is 5\% in default for MMHC and Greedy Search; threshold on mutual information test is 1\% in default for PC; prior type is chosen to be BDeu score\cite{Buntine91theoryrefinement} with Dirichlet Weight equals 10 for Greedy Search and MMHC. As for ARACNE, we implement the algorithm in person due to its simplicity, and the low values threshold $\tau$ are selected manually based on tradeoff between true positives and false positives, shown in Table \ref{tab:aracneParam}.

\begin{table}[htb]
\caption{\label{tab:aracneParam}low values threshold $\tau$ in ARACNE implementation for all Datasets}
\begin{center}
\begin{tabular}{|c|c|}
\hline
Dataset Name & low values threshold $\tau$ \\
\hline
alarm     & 0.01  \\
insurance & 0.05  \\
win95pts  & 0.025 \\
pigs      & 0.01  \\
link      & 0.005 \\
\hline
\end{tabular}
\end{center}
\end{table}

It's impropriate to make comparison between \emph{LSBN} and other algorithms directly for the result of \emph{LSBN} is determined by two factors: (1) the performance of Bayesian model averaging on five benchmark datasets; (2) the performance of \emph{LSBN} to divide-and-conquer Bayesian model averaging. Due to the intractability of Bayesian model averaging, we expect to evaluate the framework of \emph{LSBN} itself per se. If the performance of \emph{LSBN} is proved to be satisfactory, we would conclude that \emph{LSBN} could be well-applied to Bayesian model averaging as well.

As for the evaluation of \emph{LSBN} framework, we slightly change \emph{LSBN} to make more fair comparison between ARACNE, PC, Greedy Search and MMHC. Specifically, we replace the Bayesian model averaging process(Step 3 in Algorithm. \ref{alg:learn}) with corresponding targeted structure learning algorithm. For example, given MMHC algorithm as comparison target, we would use a modified version of \emph{LSBN} whose structure learning algorithm in LocalLearn is also MMHC while keep everything else unchanged. As a result, the performance of \emph{LSBN} framework could be measured independently, regardless of influence brought from the usage of different structure learning algorithms.

\subsection{Performance Evaluation}
\label{exp:performEval}

As for the evaluation of structure learning, we regard the Bayesian structure learning problem as a binary classification problem. For each pair of nodes, the Bayesian structure learning algorithm either assigns a positive label or a negative label to declare whether there is an edge existing between them or not.

We use \emph{precision}, \emph{recall} and \emph{F-score} as our metrics to evaluate the performances of \emph{LSBN} system. These metrics are defined as follows:

\begin{equation}
\begin{aligned}
Precision &= TP/(TP+FP)  \nonumber \\
Recall &= TP/(TP+FN)  \nonumber \\
F-Score &= \frac{2*Precision*Recall}{Precision+Recall}
\end{aligned}
\end{equation}

Where TP(True Positive) is the number of positive edges correctly classified as positive, corresponding to the hitting edges; FP(False Positive) is the number of negative edges mistakenly classified as positive, corresponding to the additional edges (error edges); TN(True Negative) is the number of negative edges correctly classified as negative; and FN(False Negative) is the number of positive edges mistakenly classified as negative, corresponding to the missing edges.

The comparison results between \emph{LSBN} and other state-of-art algorithms such as ARACNE, PC, Greedy Search and MMHC are depicted in Table \ref{tab:alarmComp} on Alarm Dataset, Table \ref{tab:insuranceComp} on Insurance Dataset, Table \ref{tab:win95ptsComp} on Win95pts Dataset, Table \ref{tab:pigsComp} on Pigs DataSet and Table \ref{tab:linkComp} on Link DataSet. The hit edge number, additional edge number and missing edge number as well as corresponding metrics such as Precision, Recall and F-Score of network results reconstructed \emph{LSBN} are shown versus their counterparts generated by the algorithms performed in global space. Note that Bayesian model averaging, abbreviated as 'Model Avg', is intractable in global scope, so the relevant blankets are filled with 'NA'.

The precision of \emph{LSBN} shows little inferiority to global in benchmark algorithms such as ARACNE, PC and MMHC, but superiority in Greedy Search (Table \ref{tab:precisonComp}). The recall of \emph{LSBN} shows superiority to global in ARACNE and PC, but little inferiority in Greedy Search and MMHC (Table \ref{tab:recallComp}). The F-Score serves as the harmonic mean of precision and recall, which shows comparable results to global in ARACNE, PC, Greedy Search and MMHC (Table \ref{tab:fscoreComp}). The results of precision, recall and F-Score reveal that \emph{LSBN} per se does not introduce noticeable errors in the procedure of partition, sampling, intra-community structure learning and mergence. What's more, in some cases, \emph{LSBN} even improve the learning quality.

As for our target, Bayesian model averaging, there is no comparison result available. By referring to other algorithms, it performs well in datasets such as Alarm, Insurance, Win95pts and link. Despite the significant disparity in Pigs, the results learned by using Bayesian model averaging is close to the results learned by other state-of-art algorithms in most cases.

\begin{table*}[htb]
\caption{\label{tab:alarmComp}Evaluation of \emph{LSBN} against other algorithms on Alarm Dataset}
\begin{center}
\begin{tabular}{|c|c|c|c|c|c|c|c|c|c|c|c|c|}
\hline
 &\multicolumn{6}{c|}{\emph{LSBN}}&\multicolumn{6}{c|}{Global}\\
\cline{2-13}
 & Hit(TP) & Miss(FN) & Error(FP) & Precision & Recall & F-Score & Hit(TP) & Miss(FN) & Error(FP) & Precision & Recall & F-Score \\
\hline
ARACNE & 39 & 7 & 6 & 86.667 & 84.783 & 85.714 & 31 & 15 & 4 & 88.571 & 67.391 & 76.543 \\
PC     & 38 & 8 & 0 & 100    & 82.609 & 90.476 & 35 & 11 & 0 & 100    & 76.087 & 86.420 \\
Greedy & 43 & 3 & 4 & 82.979 & 84.783 & 83.871 & 44 & 2  & 9 & 83.019 & 95.652 & 88.889 \\
MMHC   & 43 & 3 & 3 & 93.478 & 93.478 & 93.478 & 44 & 2  & 1 & 97.778 & 95.652 & 96.703 \\
\hline
Model Avg & 43 & 3 & 8 & 84.314 & 93.478 & 88.660 & NA & NA & NA & NA & NA & NA \\
\hline
\end{tabular}
\end{center}
\end{table*}

\begin{table*}[htb]
\caption{\label{tab:insuranceComp}Evaluation of \emph{LSBN} against other algorithms on Insurance Dataset}
\begin{center}
\begin{tabular}{|c|c|c|c|c|c|c|c|c|c|c|c|c|}
\hline
 &\multicolumn{6}{c|}{\emph{LSBN}}&\multicolumn{6}{c|}{Global}\\
\cline{2-13}
 & Hit(TP) & Miss(FN) & Error(FP) & Precision & Recall & F-Score & Hit(TP) & Miss(FN) & Error(FP) & Precision & Recall & F-Score \\
\hline
ARACNE & 33 & 19 & 4 & 89.189 & 63.462 & 74.157 & 25 & 27 & 2 & 92.593 & 48.077 & 63.291 \\
PC     & 36 & 16 & 3 & 92.308 & 69.231 & 79.121 & 31 & 21 & 1 & 96.875 & 59.615 & 73.810 \\
Greedy & 41 & 11 & 7 & 87.179 & 65.385 & 82.000 & 47 & 5  & 11& 81.034 & 90.385 & 85.455 \\
MMHC   & 43 & 9  & 4 & 91.489 & 82.692 & 86.869 & 43 & 9  & 2 & 95.556 & 82.692 & 88.660 \\
\hline
Model Avg & 45 & 7 & 7 & 86.538 & 86.538 & 86.538 & NA & NA & NA & NA & NA & NA \\
\hline
\end{tabular}
\end{center}
\end{table*}

\begin{table*}[htb]
\caption{\label{tab:win95ptsComp}Evaluation of \emph{LSBN} against other algorithms on Win95pts Dataset}
\begin{center}
\begin{tabular}{|c|c|c|c|c|c|c|c|c|c|c|c|c|}
\hline
 &\multicolumn{6}{c|}{\emph{LSBN}}&\multicolumn{6}{c|}{Global}\\
\cline{2-13}
 & Hit(TP) & Miss(FN) & Error(FP) & Precision & Recall & F-Score & Hit(TP) & Miss(FN) & Error(FP) & Precision & Recall & F-Score \\
\hline
ARACNE & 81 & 31 & 39 & 67.500 & 72.321 & 69.828 & 53 & 59 & 8 & 86.885 & 47.321 & 61.272 \\
PC     & 64 & 48 & 8  & 88.889 & 57.143 & 69.565 & 38 & 74 & 3 & 92.683 & 33.929 & 49.673 \\
Greedy & 99 & 13 & 143& 40.909 & 88.393 & 55.932 & 94 & 18 &106& 47.000 & 83.929 & 60.256 \\
MMHC   & 92 & 20 & 56 & 62.162 & 82.143 & 70.769 & 90 & 22 & 32& 73.770 & 80.357 & 76.923 \\
\hline
Model Avg & 98 & 14 & 93 & 51.309 & 87.500 & 64.686 & NA & NA & NA & NA & NA & NA \\
\hline
\end{tabular}
\end{center}
\end{table*}

\begin{table*}[htb]
\caption{\label{tab:pigsComp}Evaluation of \emph{LSBN} against other algorithms on Pigs Dataset}
\begin{center}
\begin{tabular}{|c|c|c|c|c|c|c|c|c|c|c|c|c|}
\hline
 &\multicolumn{6}{c|}{\emph{LSBN}}&\multicolumn{6}{c|}{Global}\\
\cline{2-13}
 & Hit(TP) & Miss(FN) & Error(FP) & Precision & Recall & F-Score & Hit(TP) & Miss(FN) & Error(FP) & Precision & Recall & F-Score \\
\hline
ARACNE & 592 & 0  & 14& 97.690 & 100.00 & 98.831 & 592& 15 & 4 & 99.831 & 100.00 & 99.916 \\
PC     & 574 & 18 & 0 & 100.00 & 96.959 & 98.456 & 591& 1  & 8 & 98.664 & 99.831 & 99.244 \\
Greedy & 570 & 22 & 13& 97.770 & 96.284 & 97.021 & 592& 0  & 47& 92.645 & 100.00 & 96.182 \\
MMHC   & 574 & 18 & 2 & 99.653 & 96.959 & 98.288 & 592& 0  & 0 & 100.00 & 100.00 & 100.00 \\
\hline
Model Avg & 447 & 145 & 940 & 32.228 & 75.507 & 45.174 & NA & NA & NA & NA & NA & NA \\
\hline
\end{tabular}
\end{center}
\end{table*}

\begin{table*}[htb]
\caption{\label{tab:linkComp}Evaluation of \emph{LSBN} against other algorithms on Link Dataset}
\begin{center}
\begin{tabular}{|c|c|c|c|c|c|c|c|c|c|c|c|c|}
\hline
 &\multicolumn{6}{c|}{\emph{LSBN}}&\multicolumn{6}{c|}{Global}\\
\cline{2-13}
 & Hit(TP) & Miss(FN) & Error(FP) & Precision & Recall & F-Score & Hit(TP) & Miss(FN) & Error(FP) & Precision & Recall & F-Score \\
\hline
ARACNE & 422 & 703 & 338 & 55.526 & 37.511 & 44.775 & 444& 681&280 & 61.326 & 39.467 & 48.026 \\
PC     & 466 & 659 & 277 & 62.719 & 41.422 & 49.893 & 70 &1055& 26 & 72.917 & 6.222  & 11.466 \\
Greedy & 413 & 712 & 342 & 54.702 & 36.711 & 43.936 & 783& 342&1374& 36.300 & 69.600 & 47.715 \\
MMHC   & 474 & 651 & 321 & 59.623 & 42.133 & 49.375 & 621& 504&418 & 59.769 & 55.200 & 57.394 \\
\hline
Model Avg & 408 & 717 & 413 & 49.695 & 36.267 & 41.932 & NA & NA & NA & NA & NA & NA \\
\hline
\end{tabular}
\end{center}
\end{table*}

\begin{table}[htb]
\caption{\label{tab:precisonComp}Significance of \emph{LSBN}'Precision normalized by the precision of global situation for four state-of-art structure learning algorithms.}
\begin{center}
\begin{tabular}{|c|cccc|}
\hline
Dataset   & ARACNE & PC & Greedy & MMHC \\
\hline
alarm     & 97.850\% & 100\%    & 99.952\% & 95.602\% \\
insurance & 96.324\% & 95.286\% & 107.583\%& 95.744\% \\
win95pts  & 77.689\% & 95.906\% & 87.040\% & 84.265\% \\
pigs      & 97.855\% & 101.354\%& 105.532\%& 99.653\% \\
link      & 90.542\% & 86.014\% & 150.694\%& 99.756\% \\
\hline
\end{tabular}
\end{center}
\end{table}

\begin{table}[htb]
\caption{\label{tab:recallComp}Significance of \emph{LSBN}'Recall normalized by the Recall of global situation for four state-of-art structure learning algorithms.}
\begin{center}
\begin{tabular}{|c|cccc|}
\hline
Dataset   & ARACNE & PC & Greedy & MMHC \\
\hline
alarm     & 125.808\%& 108.572\%& 88.637\% & 97.727\% \\
insurance & 132.001\%& 116.130\%& 72.341\% & 100\%    \\
win95pts  & 152.831\%& 168.419\%& 105.319\%& 102.223\%\\
pigs      & 100\%    & 97.123\% & 96.284\% & 96.959\% \\
link      & 95.044\% & 665.734\%& 52.746\% & 76.328\% \\
\hline
\end{tabular}
\end{center}
\end{table}

\begin{table}[htb]
\caption{\label{tab:fscoreComp}Significance of \emph{LSBN}'F-Score normalized by the F-Score of global situation for four state-of-art structure learning algorithms.}
\begin{center}
\begin{tabular}{|c|cccc|}
\hline
Dataset   & ARACNE & PC & Greedy & MMHC \\
\hline
alarm     & 111.982\%& 104.693\%& 94.355\% & 96.665\% \\
insurance & 117.168\%& 107.196\%& 95.957\% & 97.980\% \\
win95pts  & 113.964\%& 140.046\%& 92.824\% & 92.000\% \\
pigs      & 98.914\% & 99.206\% & 100.872\%& 98.288\% \\
link      & 93.231\% & 435.139\%& 92.080\% & 86.028\% \\
\hline
\end{tabular}
\end{center}
\end{table}

\section{Conclusion}

In this paper we present a novel framework for Bayesian structure learning using Model Averaging in large-scale networks, called \emph{LSBN}(Large-Scale Bayesian Network). In general, The framework follows the principle of divide-and-conquer by partitioning variables into multiple overlapping communities, learning intra-community structures individually and merging them together. Specifically, \emph{LSBN} first performs the partition by using a second-order partition strategy, called \emph{ROPART}, which is verified to achieve more robust results. Then \emph{LSBN} proposes a learning algorithm, named LocalLearn, to conduct sampling and structure learning within each overlapping community after the community is isolated from other variables by Markov Blanket. Finally \emph{LSBN} employs an efficient algorithm, called \emph{MERGENCE}, to merge structures of overlapping communities into a whole.

In comparison with other four state-of-art large-scale network structure learning algorithms such as ARACNE, PC, Greedy Search and MMHC, \emph{LSBN} shows comparable results in five common benchmark datasets, evaluated by precision, recall and f-score. What's more, \emph{LSBN} makes it possible to learn large-scale Bayesian structure by Model Averaging which used to be intractable.

In summary, \emph{LSBN} provides an scalable and parallel framework for the reconstruction of network structures. Besides, the complete information of overlapping communities serves as the byproduct, which could be used to mine meaningful clusters in biological networks, such as protein-protein-interaction network or gene regulatory network, as well as in social network.

\bibliographystyle{plain}
%\nocite{*}
\bibliography{citation}

\end{document}